# Fine-Tuned Large Language Models for Symptom Recognition from Spanish Clinical Text

Mai A. Shaaban[1], Abbas Akkasi[2, *], Adnan Khan[2], Majid Komeili[2], Mohammad Yaqub[1]

[1]Mohamed bin Zayed University of Artificial Intelligence, Abu Dhabi, UAE, [2]School of Computer Science, Carleton University, Ottawa, CA.

*Corresponding author: +1- 613 222 8134, E-mail: abbasakkasi@cunet.carleton.ca

## Abstract

The accurate recognition of symptoms in clinical reports is significantly important in the fields of healthcare and biomedical natural language processing. These entities serve as essential building blocks for clinical information extraction, enabling retrieval of critical medical insights from vast amounts of textual data. Furthermore, the ability to identify and categorize these entities is fundamental for developing advanced clinical decision support systems, aiding healthcare professionals in diagnosis and treatment planning. In this study, we participated in SympTEMIST – a shared task on the detection of symptoms, signs and findings in Spanish medical documents. We combine a set of large language models fine-tuned with the data released by the task's organizers.

## Introduction

Biomedical Named Entity Recognition (BioNER) has emerged as a critical component in the realm of biomedical and clinical informatics, with the primary goal of extracting meaningful entities from vast repositories of textual clinical data. Among the diverse categories of entities in the biomedical domain, recognizing the mention of symptoms, including signs and findings from clinical reports, stands out as an indispensable and profoundly significant task (1). The recognition of symptoms, signs, and findings in clinical reports holds profound implications for various facets of healthcare and research, making it an essential area of focus in the BioNER domain. Clinical narratives, often recorded in the form of electronic health records (EHR), radiology reports, pathology reports, and medical literature, are a treasure trove of information regarding the health status of patients. These reports contain a wealth of information about observed clinical phenomena, the manifestations of diseases, and diagnostic clues essential for patient care, clinical decision-making, and medical research. Key drivers for the importance of recognizing these entities include symptoms, signs, and findings that are fundamental components of diagnosing and managing a patient's health condition (2).

Timely and accurate recognition of these entities can significantly enhance clinical decision support systems, enabling healthcare providers to make more informed decisions and improve patient outcomes. However, the early identification of symptoms and signs, especially in the context of infectious diseases and outbreaks, is critical for public health surveillance. In addition, recognizing symptoms and findings empowers researchers to systematically extract structured information, supporting large-scale analyses, epidemiological studies, and the development of novel medical insights (2, 3).

The SympTEMIST (3) track is coordinated by the natural language processing (NLP) for the Biomedical Information Analysis team at the Barcelona Supercomputing Center and is endorsed by Spanish and European initiatives, including DataTools4Heart, AI4HF, BARITONE, and AI4ProfHealth. This event is scheduled to take place within the framework of the BioCreative 2023 conference. The SympTEMIST challenge encompasses three distinct subtasks, namely SymptomNER, SymptomNorm, and SymptomMultiNorm. In this study, we focused our participation exclusively on the initial subtask, which centers on the identification of symptoms within clinical reports. We addressed the challenge by combining the Language Model Models (LLMs) that had been fine-tuned using data specifically tailored to this task. The outcomes achieved may not be comparable to NER systems designed for general text domains. However, they offer valuable insights into the potential efficiency of ensemble models based on LLMs. This is especially true when these models are equipped with a diverse set of classifiers that have been fine-tuned for increased efficiency. The upcoming sections of this paper are organized as follows: we begin with a review of recent research related to the task. The subsequent section, "Materials and Methods" outlines our approach for addressing BioNER in Spanish clinical data along with the data splits, experimental setup, and hyperparameters. Following this, the "Results and Discussion" section is dedicated to analyzing the results obtained. Finally, the paper concludes with proposed future directions.

## Related Work

Recent progress in NLP has led to significant breakthroughs in BioNER. LLMs including Chat-GPT (4), BERT (5) and its derivatives (7– 10), have significantly impacted NER tasks. In (10), the authors introduced an innovative deep language model that incorporates both syntactic and semantic analysis to extract patient-reported symptoms from clinical text. The transformers-based NER model was devised to discern neurological signs and symptoms within the text, subsequently mapping them to clinical concepts (11). LLMs such as RoBERTa (12) and DeBERTa (13), have pushed the state-of-the-art in BioNER.

LLMs encounter challenges in handling non-English languages like Spanish, the fourth most spoken language, due to limited NLP resources (14). Recent years have witnessed a surge in publications related to BioNER in Spanish. Significant advancements have primarily manifested through collaborative initiatives such as the IberLEF (15) and BioNLP open shared tasks (16). However, despite this growing interest, handling the Spanish, especially in clinical settings poses numerous challenges. Spanish, with its rich inflectional morphology indicating various syntactic, semantic, and grammatical aspects (such as gender, number, etc.), presents unique complexities.

## Material and Methods

Our approach entails finetuning the LLMs, which include XLM-RL[1] and XLM-RB[2] (17), BBES[3] and BBS[4] (18), E5-B[5] and E5-L[6] (19). We also employ a simple ensemble technique called majority voting (MV). Opting for a solution among several alternatives can result in either an enhanced or deteriorated outcome of the MV technique, particularly if the majority of solutions are of lower quality.

---
[1] XLM-RL: XLM-RoBERTa Large
[2] XLM-RB: XLM-RoBERTa Base
[3] BBES: bsc-bio-ehr-es
[4] BBS: bsc-bio-es
[5] E5-B: multilingual-e5-base

XLM-RL (561M parameters) and XLM-RB (279M parameters) are multilingual masked language models, trained on a filtered CommonCrawl dataset of over two terabytes. We leverage BBS and BBES both pretrained on Spanish clinical data. BBS exclusively leverages biomedical resources and BBES incorporates both biomedical and EHR data. Extensive finetuning and evaluation on clinical NER tasks demonstrated their superior performance over other general-domain and domain-specific Spanish models. Lastly, we leverage E5-B (278M parameters) and E5-L (560M parameters) which are advanced text embedding models, trained through a contrastive approach using weak supervision. To mitigate risks associated with relying on a single model, we employ the MV technique, combining predictions of the six models to favor entity labels with the most frequent consensus across the ensemble.

**Experimental Setup**

The section provides an overview of the SympTEMIST corpus, preprocessing steps, dataset size, and model training configuration. The SympTEMIST corpus encompasses Spanish medical records, each uniquely identified by a filename. The dataset comprises a total of 744 examples. Each record is meticulously annotated, featuring an annotation ID, target label (SINTOMA), start and end positions within the text, and the corresponding text snippet. For preprocessing, we use the SpaCy tool for tokenization following the inside–outside–beginning (IOB) tagging scheme. We split the data into 95% training and 5% validation for our experiments. For testing, we received 250 examples from the SympTEMIST organizers. To fine-tune the models for submission, we utilized all 744 training examples, incorporating the 5% validation data. The hyperparameters for all six models were set as follows: a batch size of four, 70 training epochs in total, and a learning rate of 5e-05, employing a linear scheduler. In the context of our experiments, we utilized PyTorch and Transformers as the primary software modules.

## Results and Discussion

This section thoroughly analyzes validation and test data outcomes. As depicted in Figure 1, a consistent upward trend in performance was observed for the validation set as the number of epochs increased for all six models. This signifies that all models are learning and improving the ability to make accurate predictions, Notably, the XLM-RL model stood out by achieving the highest F1-score of 0.70. For evaluating on test data, we selected only three individual models out of the six models. The selection was based on the order of best-performing models on validation data, although validation data has a relatively smaller size than test data and may not fully capture the diversity inherent in the test data.

---

[6] E5-L: multilingual-e5-large

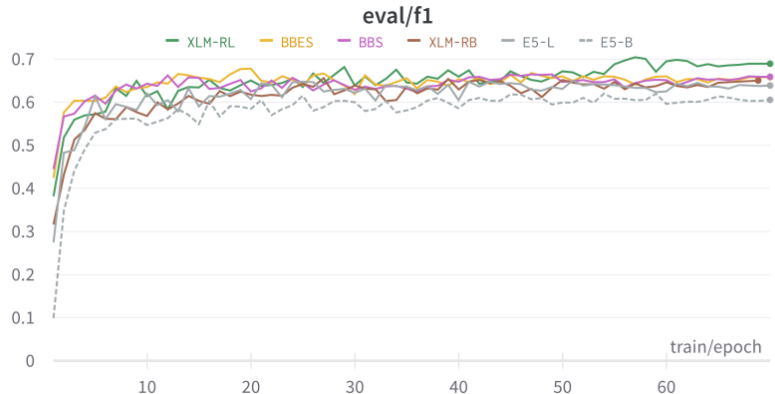

Figure 1: The f1-score trend over train epochs for validation data

Table 1 indicates that the fine-tuned models predict a few false positives, but they miss more true positive entities on test data. Additionally, if the ensemble of models in the majority voting method generates diverse predictions, it can lead to some level of vote dilution. In other words, the majority voting may average out correct predictions with incorrect ones, resulting in a less accurate overall prediction. To mitigate this problem, a weighted majority voting could be considered, where higher weights are assigned to top-performing models and lower weights to lower-performing models. Moreover, the observed results on test data with domain-specific models on Spanish clinical data highlight the pressing necessity for ongoing research and advancement in this area, as they outperform general-domain models.

| Model Variant | Validation data | | | Test data | | |
|---|---|---|---|---|---|---|
| | Precision | Recall | F1-score | Precision | Recall | F1-score |
| BBES | 0.66 | 0.70 | 0.68 | **0.70** | 0.57 | 0.63 |
| BBS | 0.63 | 0.70 | 0.66 | 0.69 | **0.62** | **0.65** |
| XLM-RL | **0.70** | **0.71** | **0.70** | 0.62 | 0.50 | 0.56 |
| MV | 0.65 | 0.62 | 0.64 | 0.68 | 0.60 | 0.64 |

Table 2: Models performance on the validation and test data

## Conclusion

In summary, the accurate identification of symptoms in clinical reports is crucial for healthcare and biomedical natural language processing. These entities are foundational for extracting vital medical insights from extensive text data. Recognizing and categorizing these entities is key for advanced clinical decision support systems to assist healthcare professionals. Our study, using domain-specific models fine-tuned with SympTEMIST data, demonstrates the urgency for continued research and progress in this field, as these models outperform general-domain ones when applied to Spanish clinical data. One potential future direction in this field involves augmenting the pool of LLMs with diverse architectures that have undergone extensive fine-tuning using substantial, similar datasets. The combination of their outcomes, taking into account their respective performance levels, is also worth considering.